\documentclass{article}

\usepackage{arxiv}

\usepackage[utf8]{inputenc} 
\usepackage[T1]{fontenc}    
\usepackage{hyperref}       
\usepackage{url}            
\usepackage{booktabs}       
\usepackage{amsfonts}       
\usepackage{nicefrac}       
\usepackage{microtype}      
\usepackage{lipsum}
\usepackage{graphicx}
\usepackage{multirow}
\usepackage{algorithm2e}
\usepackage{algorithmicx}
\usepackage{algpseudocode}
\usepackage{array}
\usepackage{mdwmath}
\usepackage{mdwtab}
\usepackage{eqparbox}
\usepackage{makecell}
\usepackage{refcount}
\usepackage{amsmath,amssymb,amsfonts}
\usepackage{graphicx,color}
\usepackage{textcomp}
\usepackage{multirow}
\graphicspath{ {./images/} }

\title{GRAD: Real-Time Gated Recurrent Anomaly Detection in Autonomous Vehicle Sensors Using
Reinforced EMA and Multi-Stage Sliding Window Techniques}

\author{
 MohammadHossein Jafari Naeimi, Ali Norouzi, Athena Abdi  \\
  Faculty of Computer Engineering, \\
K.N.Toosi University of Technology, \\
Tehran, Iran \\
  \texttt{a\_abdi@kntu.ac.ir} \\
}

\begin{document}
\maketitle
\begin{abstract}
This paper introduces GRAD, a real-time anomaly detection method for autonomous vehicle sensors that integrates statistical analysis and deep learning to ensure the reliability of sensor data. The proposed approach combines the Reinforced Exponential Moving Average (REMA), which adapts smoothing factors and thresholding for outlier detection, with the Multi-Stage Sliding Window (MS-SW) technique for capturing both short- and long-term patterns. These features are processed using a lightweight Gated Recurrent Unit (GRU) model, which detects and classifies anomalies based on bias types, while a recovery module restores damaged sensor data to ensure continuous system operation. GRAD has a lightweight architecture consisting of two layers of GRU with limited number of neurons that make it appropriate for real-time applications while maintaining the high detection accuracy. The GRAD framework achieved remarkable performance in anomaly detection and classification. The model demonstrated an overall F1-score of 97.6\% for abnormal data and 99.4\% for normal data, signifying its high accuracy in distinguishing between normal and anomalous sensor data. Regarding the anomaly classification, GRAD successfully categorized different anomaly types with high precision, enabling the recovery module to accurately restore damaged sensor data. Relative to analogous studies, GRAD surpasses current models by attaining a balance between elevated detection accuracy and diminished computational expense. These results demonstrate GRAD’s potential as a reliable and efficient solution for real-time anomaly detection in autonomous vehicle systems, guaranteeing safe vehicle operation with minimal computational overhead.
\end{abstract}


\section{Introduction}
{A}{utonomous} vehicles have recently emerged as one of the most transformative innovations in the transportation industry. Their simplicity has led to widespread adoption, offering a promising solution to many challenges associated with traditional vehicles\cite{Fang2024,Eskandarian2020}. Since a significant portion of road accidents and fatalities is primarily caused by human errors, the advancement of autonomous vehicle technology plays a key role in addressing these issues\cite{Eskandarian2020, Franco2020}. As a result, ensuring their safe and efficient operation is essential for the continued growth and development of this industry\cite{Baccari2024}.

Based on their degree of autonomy, vehicles are classified into six levels, ranging from completely non-autonomous to entirely autonomous\cite{Baccari2024}. As the autonomous system assumes greater authority over the vehicle, the importance of safety, flawless performance, and rapid response increases, and any failure in the system brings potential risks. These systems function like a layered structure, with each layer carrying out a specific task. This layered structure generally includes stages of data acquisition, data processing, and decision-making\cite{Baccari2024}. The first layer, referred to as the \textit{sensing layer}, is tasked with collecting data from the surrounding environment. In non-autonomous vehicles, this process is carried out by human senses, but in autonomous vehicles, this responsibility is taken on by a range of different sensors, each of which collects specific data from the environment\cite{Eskandarian2020}. This data may include continuous or discrete numerical values or images, which are used to simulate the function of human senses for evaluating the surrounding environment\cite{Fang2024,Bogdoll2022}.

\begin{figure}[h!]
    \centering
    \includegraphics[width=0.87\textwidth]{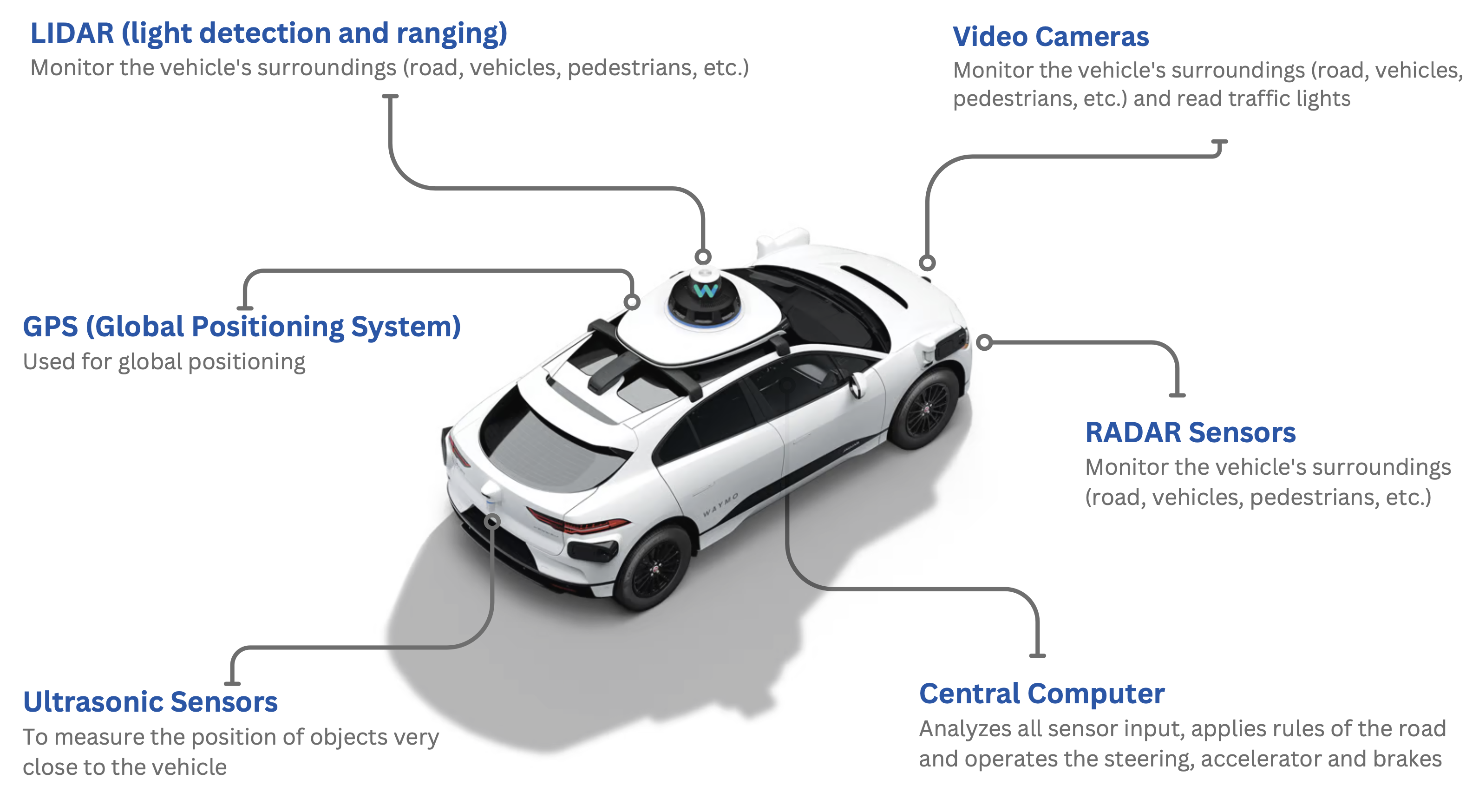}
    \caption{Illustration of the Waymo autonomous vehicle with hypothetical positions of its main sensors. The actual sensor placements may vary in real-world.}
    \label{fig:sensors-info-and-positioning}
\end{figure}

These sensors have different but complementary roles. GPS is responsible for determining the vehicle's precise geographical location and provides information such as latitude and longitude\cite{Cahyadi2023,Papadopoulos2024}. LIDAR and RADAR are used to detect surrounding objects and obstacles, create accurate 3D maps, and measure the distance and speed of objects\cite{Fang2024,Conejo2025}. Cameras, in addition to recognizing traffic signs, are used to detect road lanes, objects, and pedestrians. Ultrasonic sensors are particularly useful for short-range detection of nearby obstacles, such as parked cars or walls. The IMU measures the vehicle's acceleration and angular velocity, assisting in motion analysis\cite{Cahyadi2023,Fang2024}. Together, these sensors collaboratively collect comprehensive information about the vehicle's surroundings, enabling the autonomous system to make safe and accurate decisions\cite{Alizadeh2023}. Figure \ref{fig:sensors-info-and-positioning} shows the positioning and roles of these sensors on the vehicle.

The performance of these sensors has improved, and they are highly accurate today. Nonetheless, various factors may induce errors, resulting in undesirable system performance \cite{Eskandarian2020}. These factors include environmental conditions, such as adverse weather or unsuitable paths \cite{Kanapram2020, Yoo2025}. Sensor malfunctions can also contribute to errors, arising from hardware failures, equipment degradation, calibration issues, or power supply problems \cite{Yoo2025}. Additionally, cyber attacks pose a significant risk by tampering with data transmitted by sensors or creating interference in communications \cite{Yoo2025, Korium2024, Islam2023}.

These factors can cause changes or delays in the data transmitted from the sensors to the central unit. Generally, these changes may manifest as noise or shifts in numerical data.

Given the inability of sensors to detect and differentiate such data, inaccurate information can lead to erroneous decision-making. Data processing layer is responsible for the primary processing of raw data extracted from the sensors, and the importance of this process is considered one of the critical aspects of systems. These processes must possess high reliability and rapid response, as any delay or lack of confidence can lead to serious problems. Therefore, this layer ensures the complete accuracy of the data, detects faulty data, and attempts to replace or reconstruct it, relaying environmental information to the vehicle's control system.

Recent advancements in anomaly detection for autonomous systems have investigated diverse methodologies, encompassing statistical, machine learning, and hybrid approaches \cite{Baccari2024,Islam2023}. Statistical methods are commonly valued for their computational efficiency and capacity to analyze large datasets \cite{Carter2012,Huang2009}. Nevertheless, they often encounter difficulties in identifying intricate, dynamic anomalies \cite{Siffer2017}. On the other hand, machine learning techniques, such as autoencoders and Generative Adversarial Networks (GANs), have demonstrated significant potential in identifying anomalies \cite{Rezaei2024,Alsaade2023}. Autoencoders work by compressing and reconstructing data, while GANs generate synthetic data to resemble real data, with a discriminator identifying anomalies by distinguishing between the two \cite{Li2019,Bashar2020}. Hybrid methods integrate various techniques, leveraging the advantages of each to address the limitations \cite{Nagamani2024}. Integrating convolutional neural networks (CNNs) with LSTM models facilitates the detection of patterns, thereby enhancing the accuracy of anomaly detection \cite{Wang2023,Rajendar2022,Javed2021}. These techniques enhance the efficiency of anomaly detection systems, especially in relation to autonomous vehicles \cite{Oucheikh2020,Singh2025}.

The proposed GRAD approach leverages a hybrid anomaly detection framework that integrates statistical analysis with deep learning. Initially, raw GPS sensor data is subjected to preprocessing, during which missing values are addressed. The Reinforced Exponential Moving Average (REMA) dynamically adjusts smoothing factors and identifies sudden deviations, whereas the Multi-Stage Sliding Window (MS-SW) extracts key statistical and regression-based features to capture both short- and long-term patterns. These features are fed into a lightweight Gated Recurrent Unit (GRU) network that detects and classifies anomalies based on bias types. Furthermore, a recovery module restores damaged sensor data to ensure system reliability. The primary contributions of the proposed GRAD are delineated as follows:
\begin{itemize}
\item Presenting REMA: A dynamic statistical anomaly detection method that adapts thresholding and smoothing parameters for real-time outlier detection.
\item Multi-Stage Sliding Window (MS-SW): Deriving both statistical features (mean, variance, range) and regression-based metrics (slope, intercept, standard error) for comprehensive anomaly characterization.
\item Introducing a lightweight GRU model: Enhanced for real-time anomaly detection and classification while minimizing computational overhead.
\item Recovery module for data rectification: Reconstructing inaccurate sensor readings using EMA-based estimations to ensure continuous and accurate system operation.
\end{itemize}

The paper is organized as follows: Section II reviews related studies, Section III details the preliminaries, Section IV presents the proposed methodology, Section V discusses experimental results and evaluations, and Section VI concludes with prospective research directions.

\section{RELATED STUDIES}
The precision of the output data from these sensors in an autonomous vehicle is essential. In light of this importance, comprehensive research has been conducted to identify various types of anomalies in both structural and temporal aspects. A significant portion of these methods has been tested on time-series data for fault detection in IoT systems, whereas others have been utilized on the output data from autonomous vehicle sensors\cite{Taslimasa2023,Islam2023,Baccari2024}. These detection techniques can be classified into several categories\cite{Ane2021,Zhou2024}. These faults are categorized based on their impact and severity into three main types: point, sequence, and permanent anomalies\cite{Belay2023}. Their identification generally depends on similarity analysis or identifying inconsistencies, which can be performed via supervised or unsupervised learning methods\cite{Belay2023}.

In unsupervised learning methods, statistical approaches can be used to measure discrepancies among data points through mathematical relationships like variance or mean\cite{Carter2012,Huang2009}. These methods detect anomalies by either assigning a probability to each data point or establishing a minimum threshold for dissimilarity\cite{Singh2025,Siffer2017,Wang2020}. The widespread applicability of these methods stems from their minimal computational expense, rendering them valuable. Nonetheless, due to their significant reliance on input data, their performance tends to be relatively weaker when dealing with more complex anomalies such as intermittent or permanent anomalies compared to learning-based approaches\cite{Fang2024,Baccari2024}.

The second approach employs methods that utilize autoencoders, which feature an encoder-decoder structure to compress data into a lower-dimensional space and subsequently reconstruct it\cite{Esmaeili2023,Rajapaksha2023}. When the input data is anomalous, the model inaccurately reconstructs it, leading to a significant reconstruction error, which is then used to detect inconsistencies\cite{Shrestha2024,Alsaade2023,Rezaei2024}. However, in certain instances, they unintentionally acquire the skill to effectively reconstruct anomalies, thereby diminishing their capacity to differentiate them from normal data.

An alternative method employs Generative Adversarial Networks (GANs), which consist of two competing networks: a generator and a discriminator\cite{Bashar2020}. The generator attempts to produce data that resembles real data, whereas the discriminator seeks to differentiate between genuine and counterfeit data\cite{Li2019,Rezaei2024}. In anomaly detection, when the discriminator identifies an input sample as an outlier, it is regarded as an anomaly. GAN-based anomaly detection may experience mode collapse or instability issues, which can compromise the quality of generated data and the performance of anomaly detection\cite{Lu2023}.

Supervised learning models, especially machine learning techniques, offer another approach to classifying anomalies based on data similarity\cite{Wei2025}. These models learn patterns from labeled datasets, making them well-suited for structured anomaly detection tasks\cite{Liu2021,Li2025}. The advantage of these models is their quick detection time when analyzing real data. However, they tend to have lower accuracy, as any potential issues in the learning process can introduce bias, resulting in incorrect classification and misinterpretation during anomaly detection.

A further category of supervised models encompasses deep learning methodologies, which leverage neural network architectures to achieve highly accurate anomaly detection across various time conditions\cite{Wang2023,Su2019,Rajendar2022}. These methods utilize temporal patterns, dissimilarity detection, and threshold-based classification to effectively identify anomalies\cite{Hundman2018,Oucheikh2020,Engy2023}. However, applying deep learning methods without accounting for system constraints may result in computational burden, causing delays in detection. The computational expense of deep learning models is markedly greater than that of all previously discussed methods.

Hybrid methods that integrate various models have been proposed to tackle this challenge\cite{Nagamani2024,Sivapalan2022}. Although these methods prioritize enhancing detection efficacy, their specialized characteristics render them lacking in general applicability. Consequently, the need for a real-time and highly reliable anomaly detection system—particularly in the sensors of autonomous vehicles where both speed and accuracy are critical—is imperative.

\section{PRELIMINARIES}

\subsection{GPS Sensor and its Role in Vehicels}
\underline{G}lobal \underline{P}ositioning \underline{S}ystem (GPS) sensors are key components in modern vehicles, delivering real-time locational information via satellite signals. These sensors capture signals from a network of satellites orbiting Earth and utilize the time delay of these signals to ascertain the vehicle's position in terms of latitude and longitude. GPS sensors provide continuous tracking and are essential for numerous vehicular applications, such as navigation and route optimization\cite{Rahiman2013}.

\subsubsection{Key Parameters of GPS Data}
The primary output of a GPS sensor includes several key parameters:
\begin{itemize}
    \item \textbf{Latitude ($\phi$)}: This parameter defines the north-south position of a vehicle on the Earth’s surface. It ranges from $-90^\circ$ at the South Pole to $+90^\circ$ at the North Pole.
    \item \textbf{Longitude ($\lambda$)}: Longitude specifies the east-west position of the vehicle, with values ranging from $-180^\circ$ to $+180^\circ$.
    \item \textbf{Speed ($v$)}: This parameter measures the vehicle’s rate of motion and is typically expressed in units such as kilometers per hour (km/h) or meters per second (m/s).
\end{itemize}

\subsubsection{Triangulation and Trilateration of the GPS}
GPS operates by triangulating signals obtained from a minimum of four satellites. Each satellite broadcasts its current position and the exact time the signal was transmitted, and the vehicle's GPS receiver computes the distance to each satellite by analyzing the signal's time delay. This process, known as trilateration, involves measuring the time it takes for the signals to move from the satellites to the receiver. By multiplying the time delay by the speed of light, the receiver calculates the distance to each satellite. With distances to at least four satellites, the receiver can determine its three-dimensional position\cite{Rahiman2013,Islam2023,George2021}.

\subsubsection{Anomalies in GPS Sensors}
In autonomous vehicles, GPS allows self-localization within a map, trajectory tracking, and route decision-making\cite{Jiang2024}. Anomalies in the GPS data, including abrupt location changes, can indicate potential issues, such as sensor malfunctions or unexpected behavior. These anomalies can be categorized into several types, each representing a distinct form of deviation from the expected GPS behavior\cite{Franco2020,Ane2021}. The main types of GPS anomalies are:

\begin{itemize}
    \item \textbf{Outliers or Noises}: Outliers refer to GPS readings that deviate significantly from the expected location or movement. Sensor errors, satellite signal interference, or multipath errors can cause these outliers. Outliers typically show abrupt, large changes in location that do not correspond to the vehicle’s usual trajectory.

    \item \textbf{Jumps or Shifts}: These anomalies occur when the vehicle’s location suddenly shifts by a large distance, even though the vehicle has not physically moved that far. This can happen due to a temporary loss of GPS signals or errors in the satellite data processing. Jumps are often seen when the vehicle enters an area with weak GPS reception, like tunnels, dense urban areas, or forests, and the GPS receiver temporarily provides incorrect or inconsistent data.
\end{itemize}

\begin{figure}[h]
    \centering
    \includegraphics[width=0.77\textwidth]{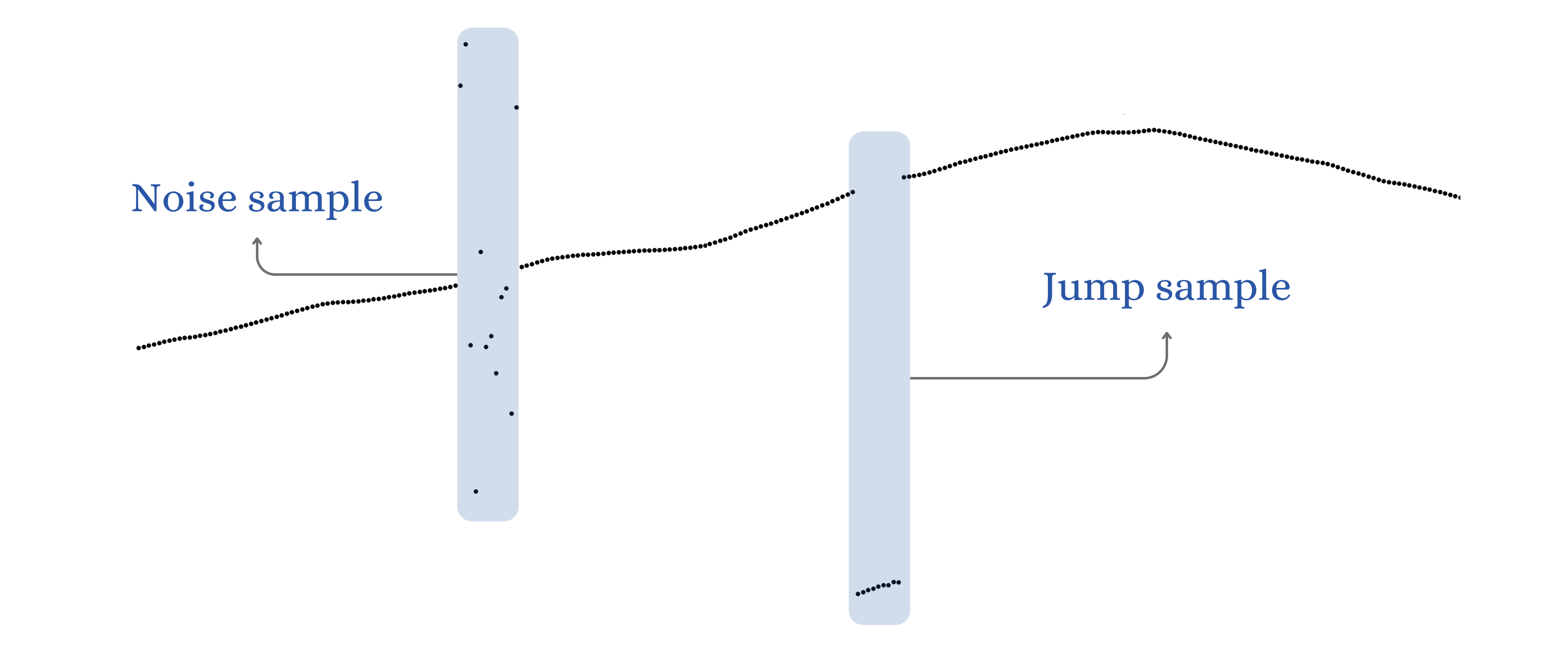}
    \caption{Types of GPS anomalies in location data of the vehicles.}
    \label{fig:jump-noise}
\end{figure}

\subsection{\underline{E}xponential \underline{M}oving \underline{A}verage (EMA)}

Moving Averages (MA) are a lightweight and simple statistical method used for estimating and predicting numerical values. A moving average is commonly used with time series data to smooth out short-term fluctuations and highlight longer-term trends or cycles. The Exponential Moving Average (EMA) is a variation of this method that, unlike traditional MA, assigns more weight to recent data points, allowing the model to respond more quickly to changes.

The formula for calculating the Exponential Moving Average (EMA) is:

\[
EMA_t = \alpha \cdot X_t + (1 - \alpha) \cdot EMA_{t-1}
\]

Where:
\begin{itemize}
    \item \( EMA_t \) is the current Exponential Moving Average value.
    \item \( X_t \) is the current data point value.
    \item \( EMA_{t-1} \) is the previous Exponential Moving Average value.
    \item \( \alpha \) is the smoothing factor.
\end{itemize}

This formula indicates that the current Exponential Moving Average (EMA) is a weighted sum of the current data point and the previous Exponential Moving Average (EMA), with more emphasis on the current value. This responsiveness makes the Exponential Moving Average (EMA) highly effective in detecting unusual changes and responding to them rapidly. By utilizing the Exponential Moving Average (EMA) formula, one can efficiently process data and identify sudden shifts.

\subsection{\underline{G}ated \underline{R}ecurrent \underline{U}nit (GRU)}
Gated Recurrent Unit networks are a variant of recurrent neural networks designed to capture sequential dependencies efficiently while mitigating the vanishing gradient problem\cite{Cho2014}. Unlike LSTM networks, GRUs simplify the architecture by combining the forget and input gates into a single update gate, leading to fewer parameters and faster training\cite{Chung2014}. This makes GRUs an effective choice for modeling sequential data such as time series, speech, and text\cite{SHERSTINSKY2020132306}. A GRU unit consists of the following components:

\begin{itemize}
    \item \textbf{Reset Gate ($r_t$)}: Determines how much past information should be ignored.
    \[
        r_t = \sigma(W_r \cdot [h_{t-1}, x_t] + b_r)
    \]

    \item \textbf{Update Gate ($z_t$)}: Controls how much of the previous hidden state should be retained and how much new information should be added.
    \[
        z_t = \sigma(W_z \cdot [h_{t-1}, x_t] + b_z)
    \]

    \item \textbf{Candidate Hidden State ($\tilde{h}_t$)}: Computes a potential new hidden state based on the current input and past information.
    \[
        \tilde{h}_t = \tanh(W_h \cdot [r_t \ast h_{t-1}, x_t] + b_h)
    \]

    \item \textbf{Hidden State ($h_t$)}: Represents the final output of the GRU unit, computed as a combination of the previous hidden state and the candidate hidden state.
    \[
        h_t = z_t \ast h_{t-1} + (1 - z_t) \ast \tilde{h}_t
    \]
\end{itemize}

\begin{figure}[h!]
    \centering
    \includegraphics[width=0.50\textwidth]{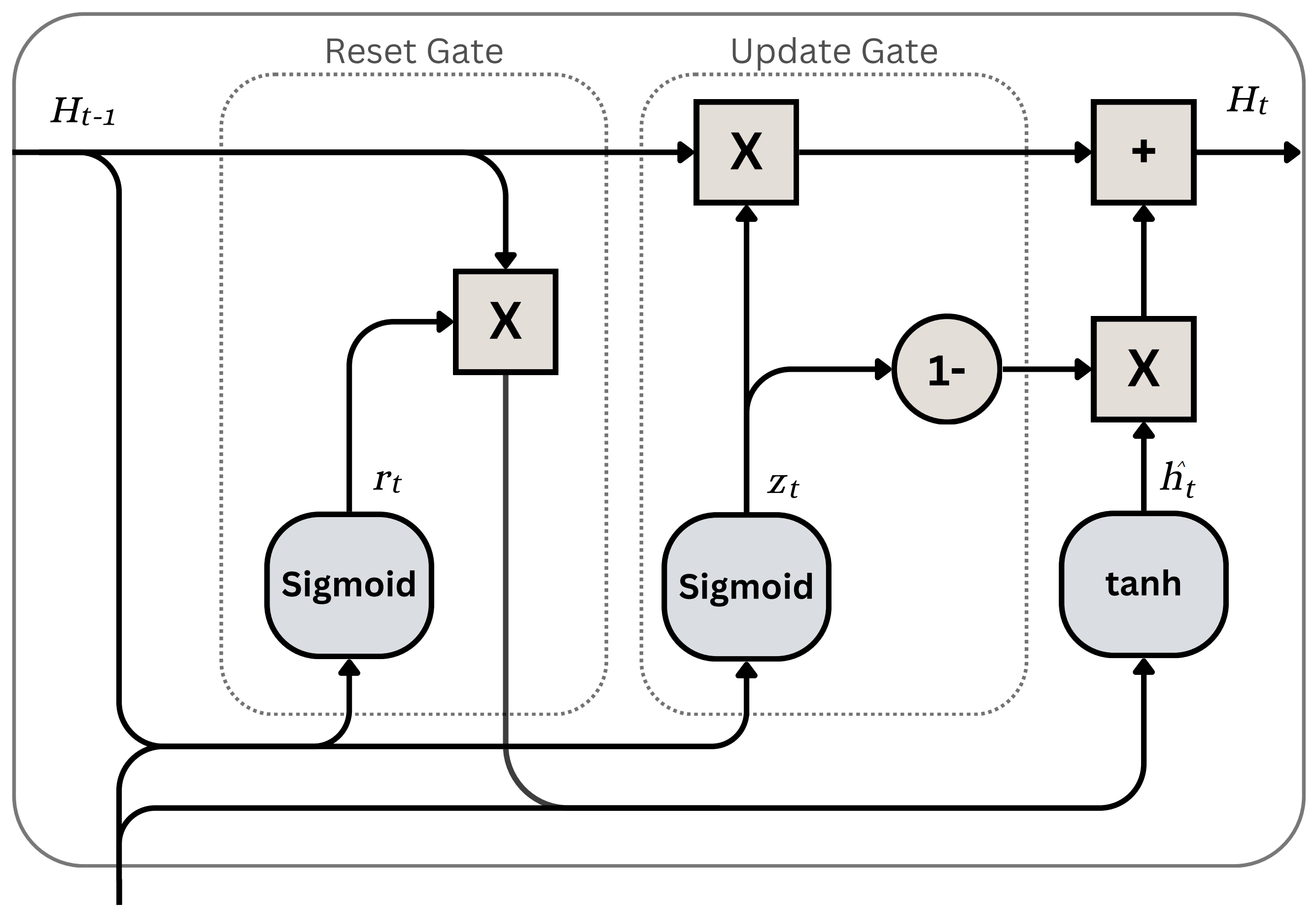}
    \caption{GRU structure highlighting its gates and operation. }
    \label{fig:GRU}
\end{figure}

Here, $\sigma$ represents the sigmoid activation function, $W$ and $b$ are the learnable weights and biases, $x_t$ is the input, and $h_{t-1}$ is the previous hidden state. The simplified structure of GRUs reduces computational complexity while preserving the ability to capture long-term dependencies in sequential data.

\subsection{Linear Regression and Standard Error}
Linear regression is a statistical method used to model the relationship between a dependent variable $y$ and one or more independent variables $x$. It is represented by the equation:
\[y = mx + c\]
where:
\begin{itemize}
    \item $m$: The slope of the line, indicating the rate of change of $y$ with respect to $x$.
    \item $c$: The intercept, representing the value of $y$ when $x = 0$.
\end{itemize}

The standard error (SE) of regression quantifies the accuracy of the linear regression model by calculating the average distance of the observed data points from the fitted regression line. It is defined as:

\[
SE = \sqrt{\frac{\sum_{i=1}^n (y_i - \hat{y}_i)^2}{n - 2}}
\]

where:
\begin{itemize}
    \item $y_i$: The observed value at point $i$.
    \item $\hat{y}_i$: The predicted value at point $i$ based on the regression line.
    \item $n$: The number of data points in the sample.
\end{itemize}

\section{Methodology}
Detecting anomalies in sensor data is essential for maintaining the reliability of autonomous systems. This study introduces a thorough methodology that combines statistical methods and deep learning to improve anomaly detection. The data preprocessing organizes sensor readings in chronological order, addresses missing values, and applies normalization to maintain consistency. A novel Reinforced Exponential Moving Average (REMA) method improves traditional EMA by dynamically adjusting its smoothing factor and integrating adaptive thresholding for effective anomaly detection. We utilize a Multi-Stage Sliding Window (MS-SW) technique to capture both short- and long-term data patterns, extracting regression-based and statistical features. The integration of REMA-based features improves resilience to noise and abrupt variations in sensor data. A Gated Recurrent Unit (GRU)-based deep learning model is then employed to learn complex temporal dependencies and categorize types of anomaly bias. The model processes sequential patterns through a stacked GRU architecture, leveraging extracted statistical and EMA features to enhance classification accuracy. A rule-based time classifier categorizes identified anomalies into transient, intermittent, or permanent types according to their temporal recurrence patterns. This classification offers enhanced understanding of system failures, enabling proactive mitigation strategies. The subsequent sections will provide a detailed examination of each key component of the proposed methodology, elucidating their individual roles in enhancing the accuracy and adaptability of anomaly detection within autonomous vehicle sensor systems.

\subsection{Preprocessing}
The sensor readings require initial preprocessing to detect missing values, establish a temporal sequence, and apply normalization. The values are initially organized according to the timeline, and simultaneous readings are merged. In the next step, missing values are identified and interpolated. Inconsistencies caused by different value ranges under different conditions can be addressed by applying a standard normalization. These measures guarantee data consistency and improve the efficacy of primary processing operations.

\subsection{Reinforced EMA (REMA)}
The Reinforced Exponential Moving Average (REMA) method enhances traditional EMA by dynamically adjusting its smoothing factor and incorporating adaptive thresholding for robust outlier detection.

\subsubsection{Optimization Params}
The REMA method optimizes model parameters to identify the ideal hyperparameters for the Exponential Moving Average (EMA) model. This process entails training the model on training data to identify the optimal parameter combination that enhances the model's capacity to detect outliers. The EMA allows for dynamic adjustment of key parameters, including the smoothing factor (alpha), its minimum and maximum limits (alpha\_min and alpha\_max), penalty and reward values for alpha adjustment (punish and reward), the window size (slide\_size), and static threshold multipliers (static\_mul). The optimization utilizes a grid search approach, wherein a spectrum of potential parameter values is tested to find the most efficacious configuration for outlier detection. Assessing each combination using a performance score derived from the model’s ability to detect both normal and abnormal (outlier) data points. The assessment is carried out using an F1 score, which balances precision and recall in identifying anomalies. The F1 scores are calculated for both positive (normal) and negative (anomaly) detections. The chosen parameters allow the EMA model to dynamically adapt its smoothing factor and thresholding strategy in response to the changing characteristics of the sensor data.

\subsubsection{Anomaly Detection and Adaptive Weight Adjustment}
For each specific time step, a new value for comparison is generated using a weighted combination of previous data and calculating a linear combination of preceding EMA values. This value can adapt to data fluctuations utilizing limits derived from the changes in EMA data over a preceding interval. It can detect abnormal changes in variation intensity. This procedure is executed in two phases: Initially, values are computed for a specific time step, and upon the introduction of new data, it is compared with the bounds to determine if it is outside the range. Upon detection of an outlier, the weight of the new data in subsequent calculations is diminished to avert performance disruption. This is achieved by reducing the alpha parameter, which controls the influence of new data on the EMA values and substitutes the EMA value with the average of recent observations, ensuring stability and preventing abrupt outliers from excessively influencing the trend estimation. Conversely, when outliers are absent, the alpha parameter is progressively augmented, enabling the model to adjust more effectively to standard data variations. This adaptive adjustment mechanism enhances the model’s resilience and sensitivity to data fluctuations. The distance metric, indicating the absolute difference between the raw sensor reading and the EMA value, is computed to quantify the deviation from the expected trend.

This multi-layered adaptation mechanism allows REMA to balance sensitivity to anomalies with resilience to normal variations, ensuring accurate and dynamic outlier detection.
\begin{figure}[h]
    \centering
    \includegraphics[width=0.77\textwidth]{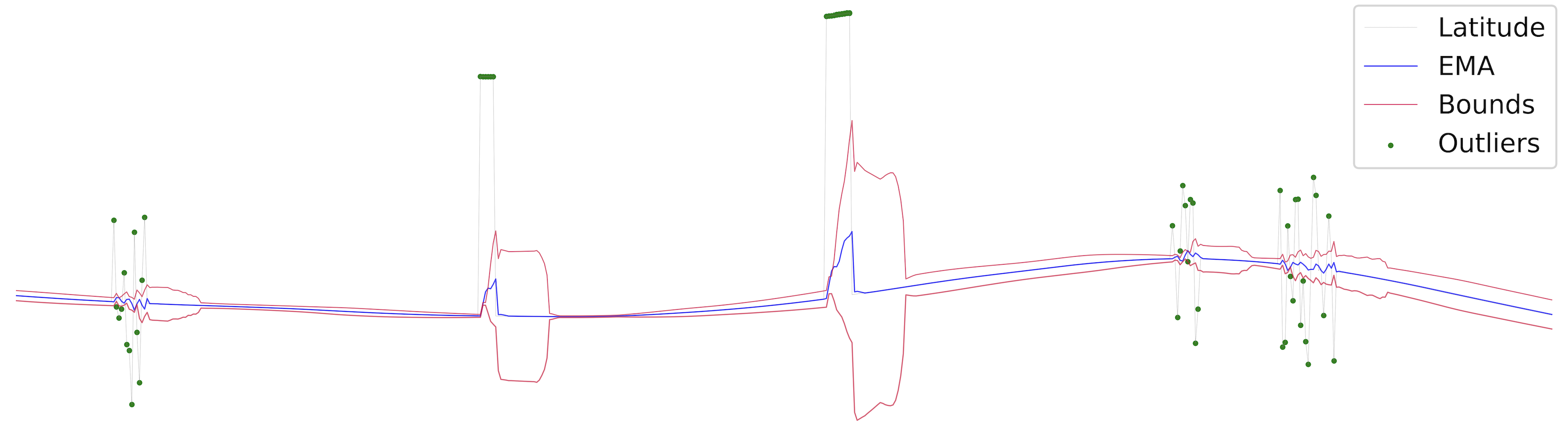}
    \caption{Illustrating REMA bounds and value adjustments.}
    \label{fig:EMA-Detection}
\end{figure}

\begin{figure*}[t]
    \centering
    \includegraphics[width=1\textwidth]{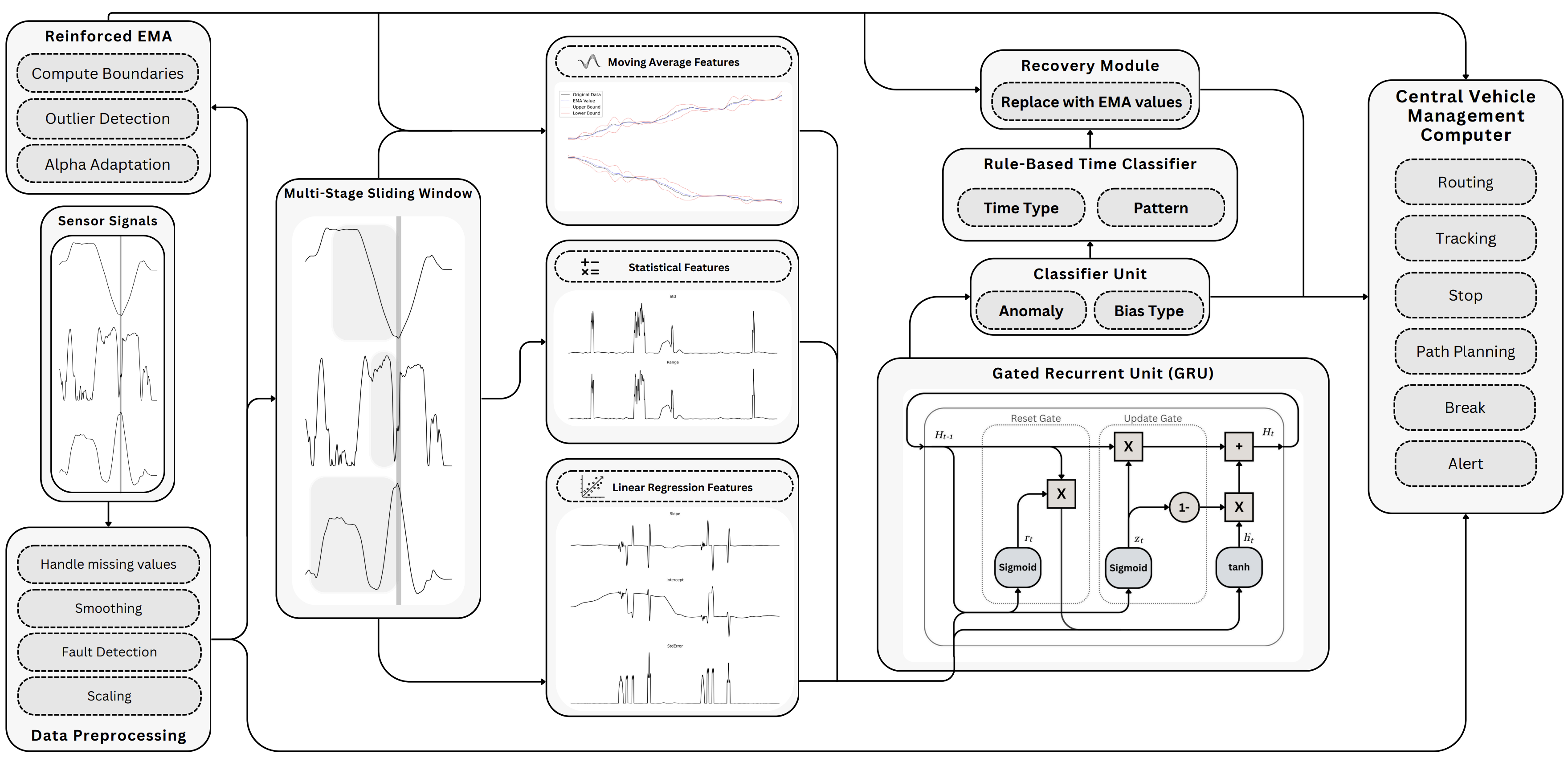} 
    \caption{GRAD Framework using REMA and GRU for anomaly detection, with a recovery module to ensure system reliability.}
    \label{fig:B-ARAD}
\end{figure*}

\begin{algorithm}
\caption{Reinforced EMA Model Functions}

\textbf{Params:} 
$a$: smoothing factor, $a_{min}$: alpha min, $a_{max}$: alpha max, $p$: punish, $r$: reward, $ss$: slide size, $S_{factor}$: sensitivity  

\vspace{0.1cm}

\textbf{Inputs:}

    $t$: current index, $row_{data}$: new data

\vspace{0.1cm}
\textbf{Outputs:}

    $y_{pred}$: Outlier predictions, $ema$ features

\SetKwFunction{Ffit}{fit}
\SetKwFunction{Fcheck}{check}

\SetKwProg{Fn}{Function}{:}{}

\vspace{0.1cm}

\Fn{\Ffit{$t$}}{

    \vspace{0.2cm}
    $p\_value$ = Mean($ema[t-ss // i ]$ for $i$ in [1, 2, 3]
    
    $ema[t]$ = $a$ * $data[t-1]$ + (1 - $a$) * $p\_value$
    
    $threshold$  = Std($ema[t - ss:t-1]$)
    
    $upper_{bound}$ = $ema[t]$ + $threshold$ * $S_{factor}$
    
    $lower_{bound}$ = $ema[t]$ - $threshold$ * $S_{factor}$

}
\vspace{0.2cm}

\Fn{\Fcheck{$t$, $row_{data}$}}{

    \vspace{0.2cm}

    \If{$row_{data}$ is outside $lower_{bound}$ or $upper_{bound}$} {

        $y_{pred}$ = 1 
        
        $ema[t]$ = Mean($ema[t - ss:t-1]$)

        $a$ = Max($a$ - $p$, $a_{min}$)
    }
    \Else{
        $a$ = Min($a$ + $r$, $a_{max}$)
    }
}

\vspace{0.1cm}

\end{algorithm}

\subsection{\underline{M}ulti-\underline{S}tage \underline{S}liding \underline{W}indow (MS-SW)}
Employing distinct processing windows for each input data enables the identification of both short-term and long-term behaviors. These processes extract key statistical features, providing information that allows the deep learning model to identify anomalies and predict their type by finding relationships among them. The subsequent sections will elucidate how each category of these features underscores a distinct type of anomaly.

Regression-based features, including slope, intercept, and standard error (SE), describe the relationship among data points over time. Fitting a regression line within each window enables the monitoring of variations in sensor data. SE measures the degree to which the regression line aligns with the data. A low SE indicates a clear trend, while a high SE suggests greater variations and uncertainty in the data. When these features are applied to normal data, SE remains low, but in noisy data, its value becomes larger than usual and effectively represents the noise behavior. Nonetheless, regression encounters difficulties with long jumps. Due to the abrupt and intense nature of this anomaly, the initial values fluctuate, but after a short period, the processing window moves over the jumped values and reports a normal trend. Still, the reported values continue to exhibit anomalous behavior, albeit with diminished accuracy.

The sliding window method utilizes a smaller window size to extract statistical features, offering a more responsive approach to rapid changes in data compared to regression-based methods. By focusing on local patterns, this technique enables a more precise correlation among data points. Key statistical features encompass Standard Deviation, which assesses data dispersion, and the Relative Strength Index (RSI), which quantifies the strength and momentum of changes within a given interval. The range captures differences between the maximum and minimum values, highlighting abrupt shifts and variation, defined as the difference between the last observed data point and the incoming data point.

The features from the previous two models accurately highlight significant values with high accuracy in noisy data. Nonetheless, their performance diminishes in identifying jump anomalies. In light of the significance of this anomaly type, we addressed this limitation by integrating REMA features. The dynamic characteristics of EMA bounds allow for shift detection, adapting by updating the alpha value to keep EMA closely aligned with the actual data. By measuring the disparity between EMA and incoming data, sudden changes can be identified, showing high deviations as potential anomalies. If EMA bands fail to detect anomalies due to excessive noise, statistical and regression features compensate by capturing underlying patterns, ensuring both types of anomalies are accurately identified. This combination tracks trends in both noisy and jumpy data. However, despite these improvements, the extracted features still contain errors, and simple statistical methods alone cannot fully distinguish anomalies. Therefore, a deep learning model with memory is essential for understanding these complex relationships.

\begin{figure}[h]
    \centering
    \includegraphics[width=0.79\textwidth]{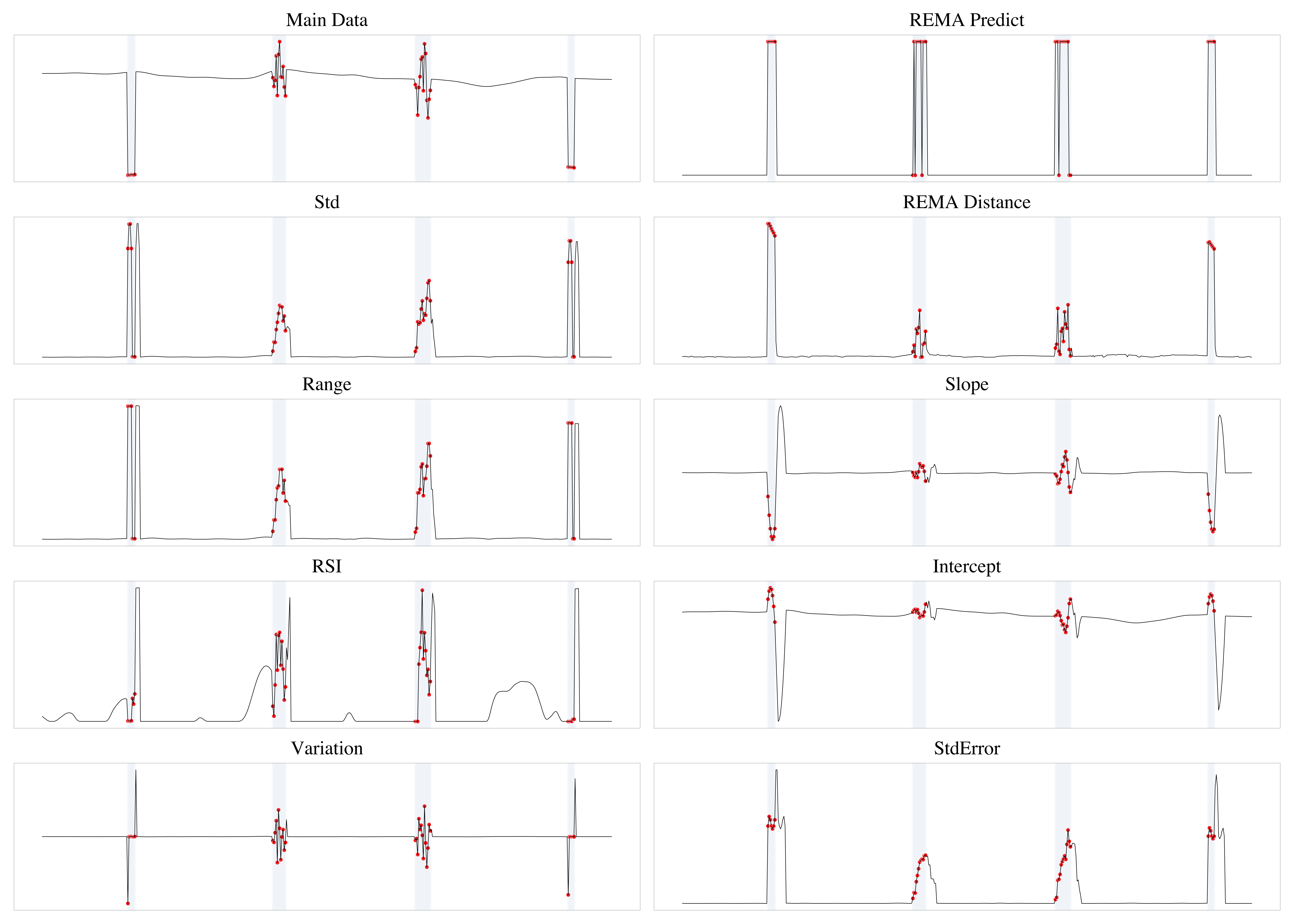}
    \label{fig:MS-SW}
    \caption{Demonstrating feature extraction using a multi-stage sliding window approach.}
\end{figure}

\subsection{Deep Learning Model (GRU)}
The deep learning model is designed to learn complex temporal dependencies in sensor data and enhance anomaly detection capabilities. A Gated Recurrent Unit (GRU) network is utilized for its proficiency in efficiently capturing sequential patterns while mitigating issues such as vanishing gradients. GRUs incorporate update and reset gates, which allow the network to selectively retain relevant information and discard redundant or irrelevant data. This selective memory mechanism guarantees the effective modeling of both short-term fluctuations and long-term dependencies. The GRU model's input comprises statistical and regression-derived features obtained via the Multi-Stage Sliding Window (MS-SW) method, along with certain features from the Reinforced EMA (REMA) model. These combined features provide a comprehensive representation of sensor behavior, enabling the network to detect subtle deviations and categorize different anomaly bias types. 

The GRU model processes sequences of extracted features and learns the underlying bias relationships that differentiate normal patterns from anomalies. As illustrated in Figure X, the extracted statistical, regression-based, and EMA features are passed to the anomaly detection model, where a stacked Gated Recurrent Unit (GRU) network processes sequential patterns to identify anomalies. The input sequences, organized by time steps, are fed into the first GRU layer, comprising 32 units. This layer utilizes update and reset gates to capture temporal dependencies, preserving pertinent information while eliminating superfluous data. The processed sequences then pass through a second GRU layer with 16 units, which further refines the extracted bias relationships, strengthening the model’s capacity to distinguish between normal patterns and anomalies. The final output is obtained through a fully connected dense layer with two output neurons, activated by a softmax function, classifying each time step as either normal or anomalous. We classified the bias types of anomalies to improve interpretability and enable focused mitigation strategies. In the subsequent phase, the identified anomalies are classified according to their temporal attributes and recurrence patterns. The output of the GRU-based detection model is transmitted to an anomaly classification module to accomplish this. 

\subsection{Rule-Based Time Classifier and Recovery}
The time categorization of anomaly types post-detection enables the formulation of an appropriate framework for assessing their impact on the system. To enhance comprehension of the various temporal categories, they may be categorized as follows.

\subsubsection{\textit{Transient}} Transient anomalies are ephemeral events that occur momentarily, often resulting from transient sensor malfunctions, environmental disturbances, or temporary fluctuations in system conditions.

\subsubsection{\textit{Intermittent}} Intermittent anomalies, in contrast, occur at irregular yet recurring intervals. These anomalies are indicative of more persistent underlying issues, such as unstable components, communication disruptions, or intermittent calibration problems. Identifying intermittent anomalies can help in pinpointing areas requiring maintenance or recalibration.

\subsubsection{\textit{Permanent}} Permanent anomalies represent long-lasting disturbances that do not resolve over time. In contrast to transient or intermittent anomalies, they can severely affect the system’s functionality. These anomalies typically signal critical failures, including sensor degradation, mechanical breakdowns, or severe misalignments.

To improve the detection and management of anomalies, a counter is utilized to track consecutive occurrences, while a memory system is implemented to identify recurring patterns, particularly in the case of intermittent anomalies. While some anomalies can be replaced, others, depending on their inherent characteristics and associated bias type, may be irreparable, requiring alternative solutions.

The recovery module replaces repairable data by detecting the anomaly time type. This process is carried out by replacing anomalous data with values computed using the EMA method. For short-term abnormal data, the EMA value provides an accurate reflection of the real data. However, for abnormal data categorized as Permanent, the model’s accuracy decreases, making these values unreliable for replacement. In such cases, the recovery module sends an alert to the main system, enabling it to take appropriate actions to handle the situation.This mechanism ensures that sensor data is processed with higher reliability and transmitted to the main system in real time.

\begin{figure}[h!]
    \centering
    \includegraphics[width=0.72\textwidth]{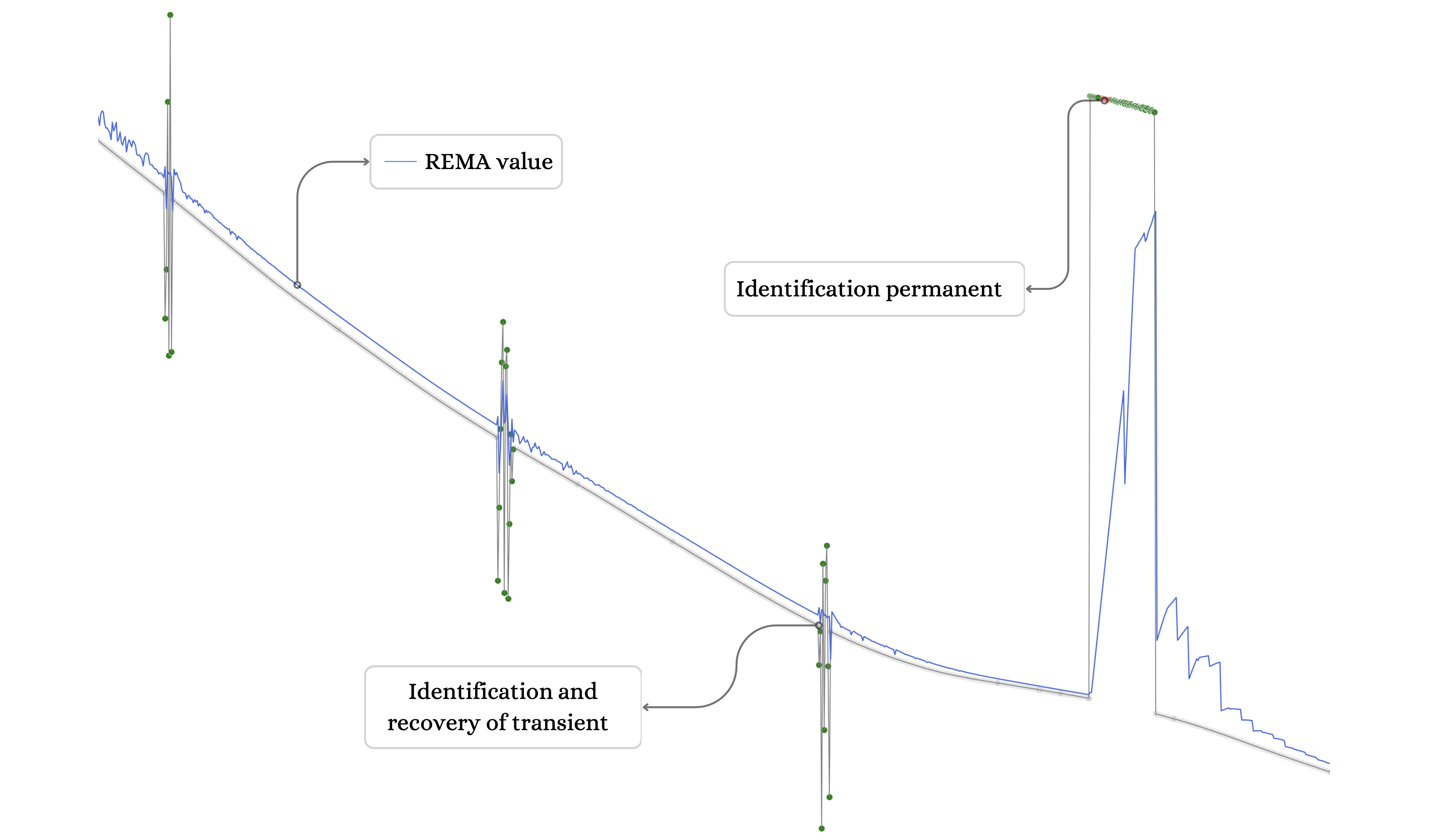}
    \caption{Illustration of EMA-Based Recovery: The EMA value adapts dynamically to smooth fluctuations and recover from anomalies.}
\end{figure}

\section{EXPERIMENTAL RESULTS}
In this section, various GPS sensor datasets are used to evaluate the proposed method. This evaluation enables a precise and real-time analysis of the model’s performance and evaluates its capabilities based on extracted features. The results are ultimately compared with prior studies to evaluate the enhancements and efficacy of the proposed method.

\subsection{Data presentation}
To assess the proposed method, it is necessary to require data based on sensor readings, thereby facilitating a controlled and systematic evaluation of anomaly detection efficacy. GPS sensor data may display numerous variations resulting from environmental influences, hardware constraints, or unforeseen interferences. By simulating different scenarios, we can analyze the model’s proficiency in detecting and classifying anomalies accurately. The assessment was performed on GPS data obtained from two datasets: MMITSS and Zurich.

\begin{enumerate}

    \item MMITSS: This GPS dataset was collected as part of the Multi-Modal Intelligent Transportation Signal Systems (MMITSS) study, which aims to enhance traffic signal optimization using advanced communication technologies. The dataset contains real-world traffic data from the Arizona Connected Vehicle Test Bed in Anthem, Arizona, covering six signalized intersections along a 1.9-mile arterial corridor. The dataset includes second-by-second vehicle trajectories, capturing longitude, latitude, altitude, speed, heading, and acceleration\cite{MMITSS2015}.

    \item Zurich: The Zurich Urban Micro Aerial Vehicle (MAV) GPS dataset provides a comprehensive collection of time-synchronized aerial data captured within urban streets at low altitudes (5–15 m). It includes high-resolution aerial images, global positioning system (GPS) and inertial measurement unit (IMU) sensor data, ground-level Google Street View images, and ground truth positioning data. This dataset is specifically designed to evaluate and benchmark GPS-based and vision-based localization techniques, particularly in GPS-denied urban environments where satellite signal accuracy is affected by obstructions\cite{Majdik2017}.

\end{enumerate}

Due to the absence of inherent anomalies in the datasets, a series of anomalies were artificially generated utilizing mathematical formulas predicated on mean and variance within a defined data range. The output of this process includes datasets with injected anomalies, comprising two types of anomalies: jump and noise.

\begin{table}[h]
    \centering
    \renewcommand{\arraystretch}{1.2}
    \caption{Percentage Comparison of Data Anomalies Across Two Datasets.}
    \label{tab:data_anomalies_comparison}

    \begin{tabular}{cccccc}
    \hline
    \multirow{2}{*}{Dataset} & \multirow{2}{*}{Time-Type} & \multicolumn{2}{c}{Features} & \multicolumn{2}{c}{Bias Type} \\ 
    \cline{3-6}
     &  & Latitude & Longitude & Noise & Jump \\ \hline
    \multirow{3}{*}{MMITSS} 
     & Transient    & 3.68     & 3.62      & 4.38     & 2.92     \\
     & Intermittent & 3.27     & 3.28      & 3.92     & 2.63     \\
     & Permanent    & 3.24     & 3.21      & 3.81     & 2.64     \\ \hline
     & \textbf{Total} & \textbf{10.19} & \textbf{10.11} & \textbf{12.11} & \textbf{8.19} \\ \hline
    \multirow{3}{*}{Zurich}  
     & Transient    & 3.81     & 3.48      & 4.34     & 2.96     \\
     & Intermittent & 3.18     & 3.33      & 3.89     & 2.62     \\
     & Permanent    & 3.42     & 3.22      & 4.00     & 2.64     \\ \hline
     & \textbf{Total} & \textbf{10.41} & \textbf{10.03} & \textbf{12.23} & \textbf{8.22} \\ \hline
    \end{tabular}
    
\end{table}

\subsection{Evaluation}

To assess the proposed model, we initially examine its anomaly detection accuracy to ascertain its efficacy in identifying abnormal behaviors in GPS data. Then, the model's efficacy in classifying distinct types of anomalies is evaluated to determine its capacity to differentiate between various anomalies, such as Jump and Noise
\cite{Eskandarian2020}.

\subsubsection{Detection}
The Reinforced Exponential Moving Average (REMA) method was utilized as the primary statistical model. This method, due to its low computational cost, attained an F1-score of 82\% for anomaly data and 96\% for normal data utilizing solely REMA bounds on the Zurich dataset. These results indicate that REMA performs well in scenarios where anomalies are simple. However, the precision of this model alone is insufficient for application in autonomous systems, as it faces limitations in detecting complex and severe anomalies.

The Gated Recurrent Unit (GRU) model was utilized to enhance detection accuracy. GRU utilizes a gating mechanism to retain important information while filtering out unnecessary data in time sequences.  This functionality enables the model to handle time-series data more effectively, resulting in reduced computational expenses and enhanced performance. The evaluation results demonstrate that this model achieved high accuracy in anomaly detection with a minimal error rate. The mean F1-score achieved was 97.6\% for abnormal data and 99.4\% for normal data. The findings in the Anomaly Detection Evaluation Table \ref{tab:dataset_performance} demonstrate that the GRU model has delivered consistent and precise performance across diverse anomaly scenarios and multiple GPS datasets.

\begin{table}[h]
    \centering
    \renewcommand{\arraystretch}{1.2}
    \caption{Performance Metrics of GRAD on MMITSS and Zurich Datasets.}
    \label{tab:dataset_performance}
    \begin{tabular}{l c l ccc}
    \hline
    \multirow{2}{*}{Dataset} &  &  & \multicolumn{3}{c}{Metrics} \\ \cline{4-6}
     & &  & Precision & Recall & F1 Score \\ \hline
    \multirow{2}{*}{\textbf{MMITSS}} & \multirow{2}{*}{\textbf{\rotatebox{90}{|}}} & Normal  & 99.58\% & 99.11\% & 99.34\% \\  
                                     &  & Anomaly & 96.57\% & 98.35\% & 97.45\% \\ \hline
    \multirow{2}{*}{\textbf{Zurich}} & \multirow{2}{*}{\textbf{\rotatebox{90}{|}}} & Normal  & 99.46\% & 99.09\% & 99.27\% \\  
                                     &  & Anomaly & 96.39\% & 97.82\% & 97.10\% \\ \hline
    \end{tabular}

\end{table}

\subsubsection{Classification}
\begin{figure}[h!]
    \centering
    \includegraphics[width=0.58\textwidth]{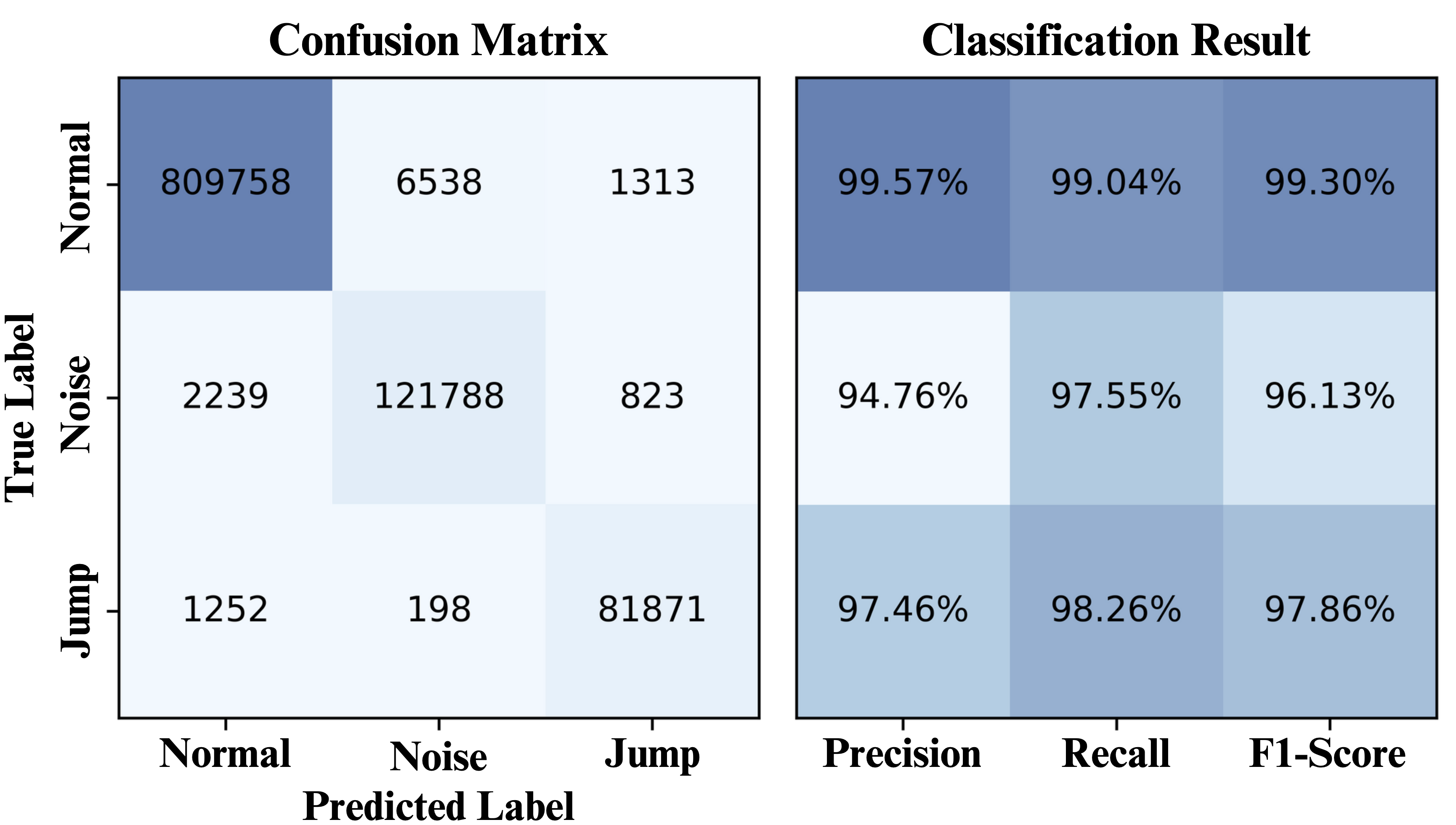}
    \caption{Confusion matrix and results of GRAD, illustrating the model’s performance in anomaly classification on the MMITSS dataset.}
    \label{fig:confusion-matrix-GRAD}
\end{figure}

Anomaly classification plays a crucial role in improving overall system performance, as accurately identifying the type of anomaly is essential for effective data recovery and correction processes. The Recovery Module leverages this classification information to detect faulty data, assess the nature of the anomaly, and substitute incorrect values with accurate ones. The Gated Recurrent Unit (GRU) model was employed for anomaly classification due to its ability to learn patterns associated with different anomaly types, enabling it to differentiate between various failure modes with considerable confidence. The results demonstrate that the model successfully categorized anomalies with high accuracy, correctly assigning them to their respective classes. Figure \ref{fig:confusion-matrix-GRAD} displays the confusion matrix and classification metrics, showcasing the performance of the GRAD model in anomaly classification.

\subsection{Analysis}
Our proposed methodology comprises several processing stages, commencing with data preprocessing and followed by extracting key features from GPS data. These features contain valuable information that allows the model to detect anomalies with reduced computational expense. In the final stage, the GRU model, which has an optimized and lightweight architecture, processes these features and generates the final predictions. This model consists of two GRU layers with 32 and 16 neurons, which significantly reduces computational cost compared to more complex models while still achieving high anomaly detection precision. 

The GRU architecture uses the feed sliding window technique to preserve temporal dependencies of prior data. This approach allows the model to make accurate predictions by analyzing the input time series data. Assessments show that with a window size of 10, the model demonstrates satisfactory performance. Nonetheless, increasing the window size to 50 enhances the model’s performance, albeit at the cost of increased computational overhead. These attributes facilitate the model's operation in real-time settings, rendering it appropriate for scenarios with constrained computational resources and the necessity for immediate detection.

\begin{figure}[h]
    \centering
    \includegraphics[width=0.48\textwidth]{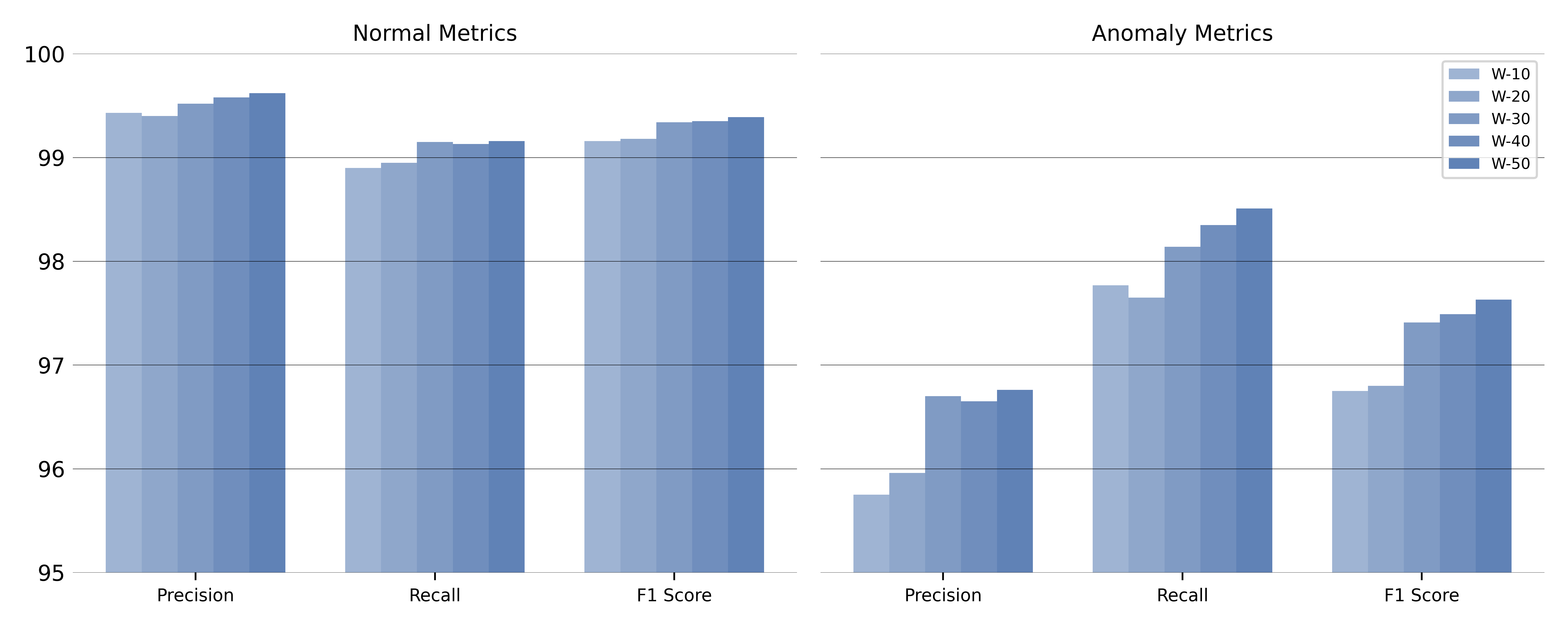}
    \caption{Performance Metrics Across Different Feed Sliding Window Sizes.}
\end{figure}

The anomaly detection model was run on a Victus 15-FA0031DX laptop, which features an Intel Core i5-12450H processor, 16 GB of RAM, and an NVIDIA GeForce GTX 1650 graphics card. The model processed data instantaneously, with each data point being processed in roughly 0.00023 seconds. This configuration facilitated efficient execution of the anomaly detection tasks, optimizing the model's performance for real-time applications.

\subsection{Comparison to Related Studies}
To evaluate the proposed model against existing methodologies, it is essential to examine the diverse techniques employed in previous studies. Some approaches primarily emphasize minimizing computational expenses; nevertheless, their methodological simplicity results in diminished accuracy in practice. These models generally employ statistical algorithms and simpler models that have lower computational costs but are unable to accurately detect anomalies in intricate and extensive datasets. Conversely, advanced techniques, including deep learning architectures and transformers, are employed for the analysis of data. These models, utilizing architectures such as autoencoders or generative transformers, can generate new data from existing data, significantly increasing computational load. Although these models provide enhanced accuracy, their computational expense is generally excessive for real-time applications. Our proposed model, which integrates lightweight techniques with an optimized GRU architecture, delivers both minimal computational expense and elevated accuracy in anomaly detection. This approach allows us to attain superior performance in real-time applications while significantly lowering computational expenses without compromising model accuracy.

\begin{table*}[!t]
    \centering
    \renewcommand{\arraystretch}{1.2}
    \small
    \caption{Evaluation of Anomaly Detection in GPS Data. Models correspond to KF, CNN, CNN-KF \cite{Franco2020}, WAVED, and MSALSTM-CNN\cite{Javed2021}.}  
    \label{tab:Compare}
    \begin{tabular}{c c c c >{\centering\arraybackslash}p{1.5cm} >{\centering\arraybackslash}p{1.5cm} >{\centering\arraybackslash}p{1.5cm} >{\centering\arraybackslash}p{1.5cm} >{\centering\arraybackslash}p{1.5cm} >{\centering\arraybackslash}p{1.5cm}}
        \hline
        \multirow{2}{*}{Injection} & \multirow{2}{*}{Magnitude} & \multirow{2}{*}{Duration} & & \multicolumn{6}{c}{Overall F1-Score \%} \\  
        \cline{5-10}  
        & & & & KF[3] & CNN[3] & CNN-KF[3] & WAVED[24] & ML-C[24] & \textbf{GRAD} \\ 
        \hline  
        

        \multirow{10}{*}{\textbf{Instant}}  & \multirow{2}{*}{25 × \textbf{$\mathcal{N}(0,0.01)$}}    & \multirow{2}{*}{1}  & \multirow{2}{*}{\textbf{\rotatebox{90}{|}}} & \multirow{2}{*}{48.5} & \multirow{2}{*}{66.0} & \multirow{2}{*}{67.4} & \multirow{2}{*}{68.9} & \multirow{2}{*}{70.2} & \multirow{2}{*}{\textbf{75.2}} \\  
                                            &                                                          &                     &                                             &                       &                       &                       &                       &                       &                                \\ \cline{2-10}
                                                          
        \multirow{2}{*}{}                   & \multirow{2}{*}{100 × \textbf{$\mathcal{N}(0,0.01)$}}   & \multirow{2}{*}{1}  & \multirow{2}{*}{\textbf{\rotatebox{90}{|}}} & \multirow{2}{*}{85.3} & \multirow{2}{*}{91.5} & \multirow{2}{*}{91.7} & \multirow{2}{*}{92.0} & \multirow{2}{*}{93.8} & \multirow{2}{*}{\textbf{97.2}} \\  
                                            &                                                          &                     &                                             &                       &                       &                       &                       &                       &                                \\  \cline{2-10}
                                                                     
        \multirow{2}{*}{}                   & \multirow{2}{*}{500 × \textbf{$\mathcal{N}(0,0.01)$}}   & \multirow{2}{*}{1}  & \multirow{2}{*}{\textbf{\rotatebox{90}{|}}} & \multirow{2}{*}{97.3} & \multirow{2}{*}{97.8} & \multirow{2}{*}{97.8} & \multirow{2}{*}{98.2} & \multirow{2}{*}{99.1} & \multirow{2}{*}{\textbf{99.5}} \\  
                                            &                                                          &                     &                                             &                       &                       &                       &                       &                       &                                \\  \cline{2-10}
                                                                     
        \multirow{2}{*}{}                   & \multirow{2}{*}{1000 × \textbf{$\mathcal{N}(0,0.01)$}}  & \multirow{2}{*}{1}  & \multirow{2}{*}{\textbf{\rotatebox{90}{|}}} & \multirow{2}{*}{98.1} & \multirow{2}{*}{98.4} & \multirow{2}{*}{98.4} & \multirow{2}{*}{98.5} & \multirow{2}{*}{98.7} & \multirow{2}{*}{\textbf{99.7}} \\ 
                                            &                                                          &                     &                                             &                       &                       &                       &                       &                       &                                \\  \cline{2-10}
                     
        \multirow{2}{*}{}                   & \multirow{2}{*}{10000 × \textbf{$\mathcal{N}(0,0.01)$}} & \multirow{2}{*}{1}  & \multirow{2}{*}{\textbf{\rotatebox{90}{|}}} & \multirow{2}{*}{99.8} & \multirow{2}{*}{99.5} & \multirow{2}{*}{99.5} & \multirow{2}{*}{99.6} & \multirow{2}{*}{99.4} & \multirow{2}{*}{\textbf{99.9}} \\  
                                            &                                                          &                     &                                             &                       &                       &                       &                       &                       &                                \\  \hline


        \multirow{10}{*}{\textbf{Constant}} & \multirow{2}{*}{\textbf{$\mathcal{U}(0,5)$}}             & \multirow{2}{*}{3}  & \multirow{2}{*}{\textbf{\rotatebox{90}{|}}} & \multirow{2}{*}{95.0} & \multirow{2}{*}{94.1} & \multirow{2}{*}{94.5} & \multirow{2}{*}{94.0} & \multirow{2}{*}{94.7} & \multirow{2}{*}{\textbf{99.5}} \\  
                                            &                                                          &                     &                                             &                       &                       &                       &                       &                       &                                \\ \cline{2-10}
                      
        \multirow{2}{*}{}                   & \multirow{2}{*}{\textbf{$\mathcal{U}(0,5)$}}             & \multirow{2}{*}{5}  & \multirow{2}{*}{\textbf{\rotatebox{90}{|}}} & \multirow{2}{*}{96.7} & \multirow{2}{*}{94.8} & \multirow{2}{*}{95.2} & \multirow{2}{*}{95.3} & \multirow{2}{*}{95.6} & \multirow{2}{*}{\textbf{99.5}} \\  
                                            &                                                          &                     &                                             &                       &                       &                       &                       &                       &                                \\  \cline{2-10}
                                 
        \multirow{2}{*}{}                   & \multirow{2}{*}{\textbf{$\mathcal{U}(0,5)$}}             & \multirow{2}{*}{10} & \multirow{2}{*}{\textbf{\rotatebox{90}{|}}} & \multirow{2}{*}{97.3} & \multirow{2}{*}{96.4} & \multirow{2}{*}{97.0} & \multirow{2}{*}{96.5} & \multirow{2}{*}{97.5} & \multirow{2}{*}{\textbf{99.3}} \\  
                                            &                                                          &                     &                                             &                       &                       &                       &                       &                       &                                \\  \cline{2-10}
                                 
        \multirow{2}{*}{}                   & \multirow{2}{*}{\textbf{$\mathcal{U}(0,3)$}}             & \multirow{2}{*}{10} & \multirow{2}{*}{\textbf{\rotatebox{90}{|}}} & \multirow{2}{*}{94.6} & \multirow{2}{*}{95.8} & \multirow{2}{*}{96.2} & \multirow{2}{*}{96.4} & \multirow{2}{*}{97.2} & \multirow{2}{*}{\textbf{99.6}} \\ 
                                            &                                                          &                     &                                             &                       &                       &                       &                       &                       &                                \\  \cline{2-10}
                    
        \multirow{2}{*}{}                   & \multirow{2}{*}{\textbf{$\mathcal{U}(0,1)$}}             & \multirow{2}{*}{10} & \multirow{2}{*}{\textbf{\rotatebox{90}{|}}} & \multirow{2}{*}{85.1} & \multirow{2}{*}{91.6} & \multirow{2}{*}{92.7} & \multirow{2}{*}{92.4} & \multirow{2}{*}{94.6} & \multirow{2}{*}{\textbf{99.7}} \\  
                                            &                                                          &                     &                                             &                       &                       &                       &                       &                       &                                \\  \hline
                    
                    
        \multirow{10}{*}{\textbf{Bias}}     & \multirow{2}{*}{\textbf{$\mathcal{U}(0,5)$}}             & \multirow{2}{*}{3}  & \multirow{2}{*}{\textbf{\rotatebox{90}{|}}} & \multirow{2}{*}{95.7} & \multirow{2}{*}{93.4} & \multirow{2}{*}{94.2} & \multirow{2}{*}{93.6} & \multirow{2}{*}{94.8} & \multirow{2}{*}{\textbf{99.4}} \\  
                                            &                                                          &                     &                                             &                       &                       &                       &                       &                       &                                \\ \cline{2-10}
                      
        \multirow{2}{*}{}                   & \multirow{2}{*}{\textbf{$\mathcal{U}(0,5)$}}             & \multirow{2}{*}{5}  & \multirow{2}{*}{\textbf{\rotatebox{90}{|}}} & \multirow{2}{*}{96.0} & \multirow{2}{*}{94.3} & \multirow{2}{*}{94.9} & \multirow{2}{*}{95.5} & \multirow{2}{*}{95.7} & \multirow{2}{*}{\textbf{99.3}} \\  
                                            &                                                          &                     &                                             &                       &                       &                       &                       &                       &                                \\  \cline{2-10}
                                
        \multirow{2}{*}{}                   & \multirow{2}{*}{\textbf{$\mathcal{U}(0,5)$}}             & \multirow{2}{*}{10} & \multirow{2}{*}{\textbf{\rotatebox{90}{|}}} & \multirow{2}{*}{96.6} & \multirow{2}{*}{95.4} & \multirow{2}{*}{96.7} & \multirow{2}{*}{95.5} & \multirow{2}{*}{97.4} & \multirow{2}{*}{\textbf{98.9}} \\  
                                            &                                                          &                     &                                             &                       &                       &                       &                       &                       &                                \\  \cline{2-10}
                                
        \multirow{2}{*}{}                   & \multirow{2}{*}{\textbf{$\mathcal{U}(0,3)$}}             & \multirow{2}{*}{10} & \multirow{2}{*}{\textbf{\rotatebox{90}{|}}} & \multirow{2}{*}{94.8} & \multirow{2}{*}{93.7} & \multirow{2}{*}{95.5} & \multirow{2}{*}{94.9} & \multirow{2}{*}{96.1} & \multirow{2}{*}{\textbf{98.9}} \\ 
                                            &                                                          &                     &                                             &                       &                       &                       &                       &                       &                                \\  \cline{2-10}
                
        \multirow{2}{*}{}                   & \multirow{2}{*}{\textbf{$\mathcal{U}(0,1)$}}             & \multirow{2}{*}{10} & \multirow{2}{*}{\textbf{\rotatebox{90}{|}}} & \multirow{2}{*}{86.9} & \multirow{2}{*}{86.8} & \multirow{2}{*}{90.0} & \multirow{2}{*}{88.1} & \multirow{2}{*}{90.6} & \multirow{2}{*}{\textbf{98.8}} \\  
                                            &                                                          &                     &                                             &                       &                       &                       &                       &                       &                                \\  \hline


        \multirow{8}{*}{\textbf{Drift}}     & \multirow{2}{*}{\textit{linspace}$\mathcal(0,4)$}        & \multirow{2}{*}{10} & \multirow{2}{*}{\textbf{\rotatebox{90}{|}}} & \multirow{2}{*}{93.4} & \multirow{2}{*}{95.5} & \multirow{2}{*}{96.1} & \multirow{2}{*}{96.2} & \multirow{2}{*}{97.5} & \multirow{2}{*}{\textbf{99.1}} \\  
                                            &                                                          &                     &                                             &                       &                       &                       &                       &                       &                                \\ \cline{2-10}
                    
        \multirow{2}{*}{}                   & \multirow{2}{*}{\textit{linspace}$\mathcal(0,4)$}        & \multirow{2}{*}{20} & \multirow{2}{*}{\textbf{\rotatebox{90}{|}}} & \multirow{2}{*}{93.8} & \multirow{2}{*}{97.2} & \multirow{2}{*}{97.4} & \multirow{2}{*}{97.6} & \multirow{2}{*}{97.7} & \multirow{2}{*}{\textbf{98.3}} \\  
                                            &                                                          &                     &                                             &                       &                       &                       &                       &                       &                                \\  \cline{2-10}
                               
        \multirow{2}{*}{}                   & \multirow{2}{*}{\textit{linspace}$\mathcal(0,2)$}        & \multirow{2}{*}{10} & \multirow{2}{*}{\textbf{\rotatebox{90}{|}}} & \multirow{2}{*}{87.9} & \multirow{2}{*}{94.0} & \multirow{2}{*}{94.2} & \multirow{2}{*}{94.9} & \multirow{2}{*}{95.6} & \multirow{2}{*}{\textbf{98.9}} \\  
                                            &                                                          &                     &                                             &                       &                       &                       &                       &                       &                                \\  \cline{2-10}
                               
        \multirow{2}{*}{}                   & \multirow{2}{*}{\textit{linspace}$\mathcal(0,2)$}        & \multirow{2}{*}{20} & \multirow{2}{*}{\textbf{\rotatebox{90}{|}}} & \multirow{2}{*}{86.7} & \multirow{2}{*}{95.3} & \multirow{2}{*}{95.9} & \multirow{2}{*}{96.1} & \multirow{2}{*}{96.1} & \multirow{2}{*}{\textbf{98.3}} \\
                                            &                                                          &                     &                                             &                       &                       &                       &                       &                       &                                \\  \hline
                    
    \end{tabular}
\end{table*}

The first approach, formulated by van Wyk et al. \cite{Franco2020}, integrates convolutional neural networks (CNNs) with Kalman filtering and a $\chi^2$-detector to detect and identify anomalies in sensor data. Their experimental results show that this hybrid approach outperforms CNNs and Kalman filtering separately, offering a high F1 score in anomaly detection. This method provides robust detection capabilities; however, the combination of CNNs with Kalman filtering significantly increases computational complexity, potentially obstructing its implementation in real-time applications with constrained computational resources.

The second method, proposed by Javed et al. \cite{Javed2021}, utilizes a combination of a multi-stage attention mechanism and a Long Short-Term Memory (LSTM)-based Convolutional Neural Network (CNN) for anomaly detection in Connected and Autonomous Vehicles (CAVs). The MSALSTM-CNN method effectively processes sensor data streams by converting them into vectors for anomaly detection. It employs a weight-adjusted fine-tuned ensemble (WAVED) approach to improve the anomaly detection rate. The method’s complexity, particularly due to the use of LSTM and CNN models, may lead to elevated computational costs, potentially restricting its real-time applicability in resource-limited environments.
Our proposed model amalgamates lightweight techniques with an optimized GRU (Gated Recurrent Unit) architecture, seeking to achieve a balance between computational efficiency and high detection accuracy. By employing a more streamlined architecture, we reduce the computational burden significantly compared to methods like MSALSTM-CNN and CNN-Kalman hybrid models.

The results from our assessment, illustrated in Table \ref{tab:Compare}, demonstrate that our model outperforms both approaches across multiple performance metrics. Our model achieves superior F1-score, precision, recall, and overall anomaly detection accuracy while also maintaining significantly lower computational costs. These findings highlight the effectiveness of our proposed model as a computationally efficient and highly accurate anomaly detection solution for autonomous vehicle sensors, establishing it as a viable alternative to current methodologies.

\section{Conclusion}
The advancement of the transportation sector and the introduction of autonomous vehicles are addressing numerous conventional transportation challenges, particularly road accidents predominantly attributed to human error. Eliminating human errors can significantly improve safety and traffic efficiency. With the proliferation of autonomous vehicles, the precision and reliability of sensor systems responsible for environment perception and real-time decision-making become increasingly important. These sensors may be influenced by environmental disturbances or technical failures, which could lead to incorrect decisions and jeopardize vehicle safety.

We propose a novel approach for anomaly detection in the sensor systems of autonomous vehicles that integrates statistical techniques with deep learning models to tackle these challenges. Our approach includes the Reinforced Exponential Moving Average (REMA) for outlier detection, a multi-stage sliding window (MS-SW) for capturing different data patterns, and a model based on the Gated Recurrent Unit (GRU) for detecting and classifying anomalous data. This technique, owing to its minimal computational expense and real-time processing ability, efficiently identifies diverse anomalies in sensor data.

Evaluation results of the REMA model show that it achieved an F1-score of 82\% for anomalous data and 96\% for normal data on the Zurich dataset. These findings demonstrate that REMA is effective in identifying basic anomalies. The model's accuracy is insufficient to recognize complex and severe anomalies, a limitation remedied by employing the GRU model. The GRU model, through its gating mechanism, has effectively retained important information and filtered out unnecessary data. This capability has enabled the model to process time-series data effectively, providing better performance than the statistical REMA-based methods. Evaluation results of the GRU model show that it achieved an average F1-score of 97.6\% for anomalous data and 99.4\% for normal data, with high accuracy and stability in anomaly detection across various GPS datasets. The model has recognized temporal dependencies in the data and enhanced information filtering.

Overall, our proposed method strikes an optimal balance between computational cost and high accuracy in anomaly detection. This functionality renders it appropriate for implementation in autonomous systems necessitating real-time detection and increased safety and reliability of autonomous vehicles. Finally, our proposed method has effectively simulated the challenges of anomaly detection in GPS sensor data, providing an efficient solution for identifying and classifying GPS anomalies in autonomous vehicle systems through the GRAD methodology.

In future endeavors, we intend to explore unsupervised learning approaches for anomaly detection, enabling the use of unlabeled datasets for more general and adaptable training. Utilizing self-supervised learning, we can augment the model’s ability to detect anomalies without relying on predefined labels. This will improve the scalability and applicability of our method across diverse and unseen scenarios in autonomous vehicle sensor data.

\bibliographystyle{unsrt}
\bibliography{GRAD}


\end{document}